\documentclass[letterpaper, 10 pt, conference]{ieeeconf} 
\IEEEoverridecommandlockouts                         
\usepackage[bookmarks=true]{hyperref}

\usepackage{graphicx}
\usepackage{wrapfig}
\usepackage{amsmath}
\usepackage{amsfonts}
\usepackage{siunitx}
\usepackage{cite}
\usepackage{cancel}
\usepackage{algorithm}
\usepackage{balance}
\usepackage{algpseudocode}

\title{\LARGE \bf Adaptive Distance Functions via Kelvin Transformation}

\author{Rafael I. Cabral Muchacho and Florian T. Pokorny
\thanks{This work was partially supported by the Wallenberg AI, Autonomous Systems and Software Program (WASP) funded by the Knut and Alice Wallenberg Foundation. The authors are with RPL, EECS, KTH Royal Institute of Technology, Stockholm, Sweden { \tt\small \{ricm, fpokorny\}@kth.se}}}

\begin{document}

\maketitle
\thispagestyle{empty}
\pagestyle{empty}

\begin{abstract}

The term \textit{safety} in robotics is often understood as a synonym for avoidance. Although this perspective has led to progress in path planning and reactive control, a generalization of this perspective is necessary to include task semantics relevant to contact-rich manipulation tasks, especially during teleoperation and to ensure the safety of learned policies. We introduce the \textit{semantics-aware distance function} and a corresponding computational method based on the Kelvin Transformation. This allows us to compute smooth distance approximations in an unbounded domain by instead solving a Laplace equation in a bounded domain. The semantics-aware distance generalizes signed distance functions by allowing the zero level set to lie inside of the object in regions where contact is allowed, effectively incorporating task semantics, such as object affordances, in an adaptive implicit representation of safe sets. In numerical experiments we show the computational viability of our method for real applications and visualize the computed function on a wrench with various semantic regions.

\end{abstract}

\IEEEpeerreviewmaketitle

\section{Introduction}

The robotics field is undergoing a shift, from hierarchical and decoupled layers for planning and control, towards integrated systems, enabling robotic manipulators to achieve reactive and intuitive behavior \cite{kappler2018real}. Methods related to artificial potential fields \cite{khatib1986real, koditschek1990robot} and control barrier functions \cite{ames2019control, marinho2019dynamic, 9682604} have been useful to encode safe reactive behavior and support the development towards integrated architectures.

A task that leverages reactive behavior is obstacle avoidance in dynamic environments. In obstacle avoidance, the object's surface regions are treated equally, so motion constraints can be integrated into lower-level control structures by using distance functions to the object. Distance functions can be efficiently queried to define safety through level-set constraints, and can also be used as repelling potential fields \cite{koptev2022neural, liu2022regularized, li2023learning, maric2024online}.

In contrast to obstacle avoidance, manipulation tasks require physical interactions with objects. Significant progress has been achieved in the field of control for manipulation tasks, both with model-based and optimization methods \cite{graesdal2024towards, heins2024force, mordatch2012contact, ereztrajectory, burgess2024reactive} and through learning-based approaches \cite{driess2022learning, sievers2022learning, rajeswaran2017learning, chi2023diffusion, zhang2023plan}.

Manipulation tasks also often involve the consideration of the semantic regions or affordances of the interacting objects, which might vary as a function of the user’s intent \cite{kokic2017affordance, wang2020affordance, 10436354, geng2023rlafford}. For instance, how one grasps a tool depends on the intention: to use a hammer, one would generally hold it by the handle while for a handover, one would grasp it by the heavier end instead. 
Also, in contrast to obstacle avoidance, manipulation tasks generally consider the task semantics through hierarchical and sequential decision-making processes.

Safety considerations in learning-based methods are relevant to robotic applications, especially when learning in real environments \cite{thumm2022provably, liu2023safe}. Although safe regions in robotic applications are usually a synonym to collision-free subsets of task configuration spaces, safety methods deal with set-based (often implicit) representations of safety, and are therefore independent of the specific task-level meaning \cite{krasowski2023provably}.

\begin{figure}
    \centering
    \includegraphics[width=\linewidth]{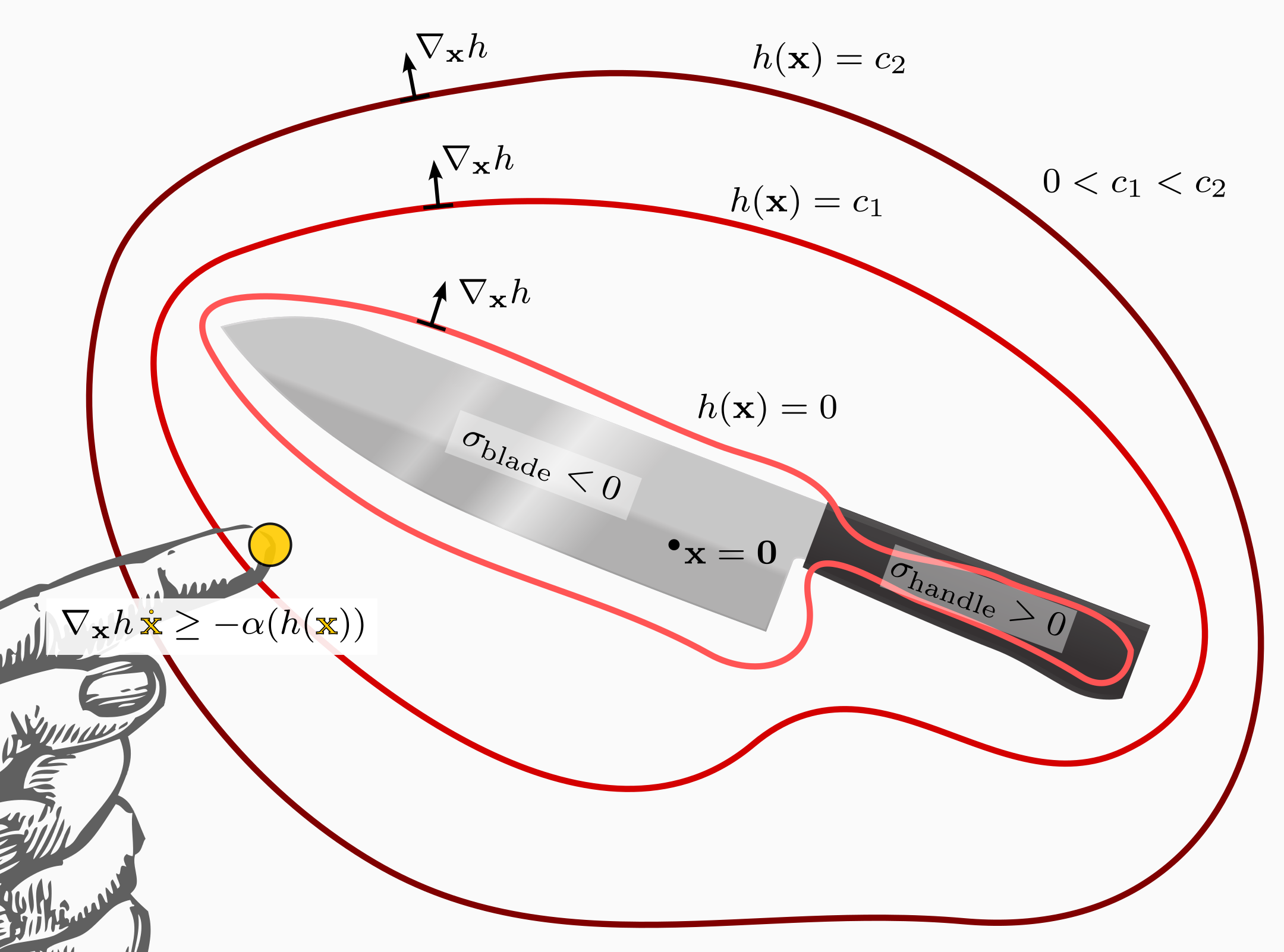}
    \vspace{-10pt}
    \caption{We implicitly define safe sets that consider the semantics of contact regions by introducing the \textit{semantics-aware distance}, $h$. Its zero level set, $h(\mathbf{x})=0$, lies within the object, where contact with the surface is allowed ($\sigma>0$). We visualize the semantics-aware distance of a knife through its iso-surfaces -- starting at the zero level set -- which describes the safe set boundary. With our proposed computation method, query times are lower than $\SI{1}{\micro\second}$ and the function is continuously differentiable and is therefore useful for example in combination with control barrier function approaches to safety critical control.}
    \label{fig:knife}
\end{figure}

The goal of this paper is to extend the applicability of safety-related methods in robotic manipulation from collision avoidance towards the consideration of object affordances and semantics while performing and learning contact-rich manipulation tasks, by introducing the concept of a \textit{semantics-aware distance}, visualized in Fig.~\ref{fig:knife}, and an efficient numerical method to compute it.

The proposed method intuitively generalizes the safety approaches for collision avoidance based on distance functions. Let a function implicitly encode the safe set as the region of the domain that maps to non-negative values. While in collision avoidance applications the zero level-set coincides with the object’s surface, we lift this constraint and treat boundary values as variables, allowing the safe-set boundary to lie within or outside of the object’s surface, depending on the time-varying or configuration-dependent semantic state of the surface region, i.e., whether contact is allowed or not.

To obtain the function, we propose a computational method that leverages recent advances in computational geometry. 
Concretely, we use the Kelvin Transformation \cite{nabizadeh2021kelvin} to solve a Boundary Value Problem defined by the Laplace equation in the exterior of an object, and discretizing the Kelvin-Transformed domain through tetrahedralization.

Two main advantages of our computational method over symbolic alternatives such as \cite{koptev2022neural, liu2022regularized, li2023learning, maric2024online} are (i) the exactness at the object's surface, matching the mesh representation even at low resolution, and (ii) faster function queries enabled by optimized tetrahedral mesh representations. Therefore, our method serves as a valuable alternative in scenarios such as industrial or workshop environments, where the tools and manipulators are generally known.  

In validation experiments shown in Section \ref{sec:validation}, we evaluate the proposed method with respect to the formalized desirables presented in Section \ref{sec:problem_statement}. We further compute the semantics-aware distance function on a combination wrench for various semantic configurations and visualize the resulting functions through their iso-surfaces. The paper concludes with comments on the limitations and applications of our approach in robotic manipulation.

\section{Preliminaries} 
\label{sec:prelim}

\subsection{Linear PDE-based Distance Field Approximations}

The homogeneous screened Poisson equation with Dirichlet boundary condition can be expressed as the boundary-value problem
\begin{align}
    u - c\Delta u = 0 \quad & \mathrm{in} \quad \Omega \notag \\
    u = u_\partial \quad & \mathrm{on} \quad \partial\Omega, \label{eq:screened_general}
\end{align}
with Laplace operator $\Delta$, screening coefficient $c \in \mathbb{R}_{\geq 0}$, boundary function $u_\partial:\partial \Omega \to \mathbb{R}$, and solution $u:\Omega \cup \partial \Omega \to \mathbb{R}$, defined in the non-empty domain $\Omega \subset \mathbb{R}^n$ and its boundary $\partial \Omega$. The Laplace equation is the special case of the homogeneous screened Poisson equation in the limit as $c\rightarrow\infty$.

The solution $u$ of the screened Poisson equation
\begin{align} \label{eq:screened_potential_field}
    u - c \Delta u = 0 \quad & \mathrm{in} \quad \Omega \notag \\
    u = 1 \quad & \mathrm{on} \quad \partial\Omega,
\end{align}
is positive and differentiable in the domain interior, and has a maximum at the boundary. Using the negative logarithm transformation
\begin{align} \label{eq:transformation_varadhan}
    w = -\sqrt{c} \,\mathrm{ln}(u),
\end{align}
the solution $u$ can be equivalently described as the solution $w$ to the regularized Eikonal equation
\begin{align} \label{eq:regularized_eikonal}
    (1 - |\nabla w|^2) + \sqrt{c} \Delta w = 0 \quad & \mathrm{in} \quad \Omega \notag \\
    w = 0 \quad & \mathrm{on} \quad \partial \Omega.
\end{align}
As $c\rightarrow 0$, the solution $w$ approaches the exact distance-to-boundary function \cite{belyaev2015variational, gurumoorthy2009schrodinger, crane2013geodesics}.

In bounded low-dimensional domains, the equation can be effectively solved through numerical approaches, i.e., finite element and finite difference methods. One strategy to solve it in exterior or unbounded domains is the Kelvin Transform which leverages an inversion map to represent the unbounded domain by a bounded domain \cite{nabizadeh2021kelvin}.

\subsection{Kelvin Transformation for PDEs on Unbounded Domains} \label{sec:prelim_kelvin}

To solve PDEs on unbounded domains, such as object exteriors, the \textit{coordinate-stretching} method was introduced in the 1970s. It treats points at infinity by transforming the unbounded domain into a finite region~\cite{grosch1977numerical}.
The Kelvin Transformation approach \cite{nabizadeh2021kelvin} fundamentally extends the coordinate-stretching approach by proposing to use the inversion map
\begin{align}
    &\Phi(\mathbf{y}) = \frac{\mathbf{y}}{|\mathbf{y}|^2}, \quad \mathbf{y} \in \Omega_\mathrm{inv} \subset \mathbb{R}^n \setminus {\mathbf{0}} \notag \\
    &\Phi^{-1}(\mathbf{x}) = \frac{\mathbf{x}}{|\mathbf{x}|^2}, \quad \mathbf{x} \in \Omega \subset \mathbb{R}^n \setminus {\mathbf{0}},\label{eq:inversion_map}
\end{align}
where $\Omega_\mathrm{inv} = \Phi^{-1}(\Omega)$, and decompose the solution $u(\mathbf{x})$ (or $U(\mathbf{y})$) into
\begin{align}
    u(\mathbf{x}) = g(\mathbf{x})v(\mathbf{x}) \quad \mathrm{or} \quad U(\mathbf{y}) = G(\mathbf{y})V(\mathbf{y}), \label{eq:kelvin_factorization}
\end{align}
where
\begin{enumerate}
    \item the analytical behavior function $g(\mathbf{x})$ or its inverted counterpart $G(\mathbf{y})=g(\Phi(\mathbf{y}))$, captures the asymptotic behavior of the solution $u$ near infinity, and
    \item the factor $v(\mathbf{x})$, or $V(\mathbf{y}) = v(\Phi(\mathbf{y}))$, is to be solved numerically. 
\end{enumerate}

The change of variables from $U$ to $V$ removes the singularity at the inversion origin -- otherwise present in coordinate stretching -- such that the resulting PDE in the inverse domain extends to the compactified domain $\Omega_\mathrm{inv} \cup \{ \mathbf{0} \}$ \cite{nabizadeh2021kelvin}.

\section{Problem Statement} 
\label{sec:problem_statement}

\begin{figure}
    \centering
    \includegraphics[width=\linewidth]{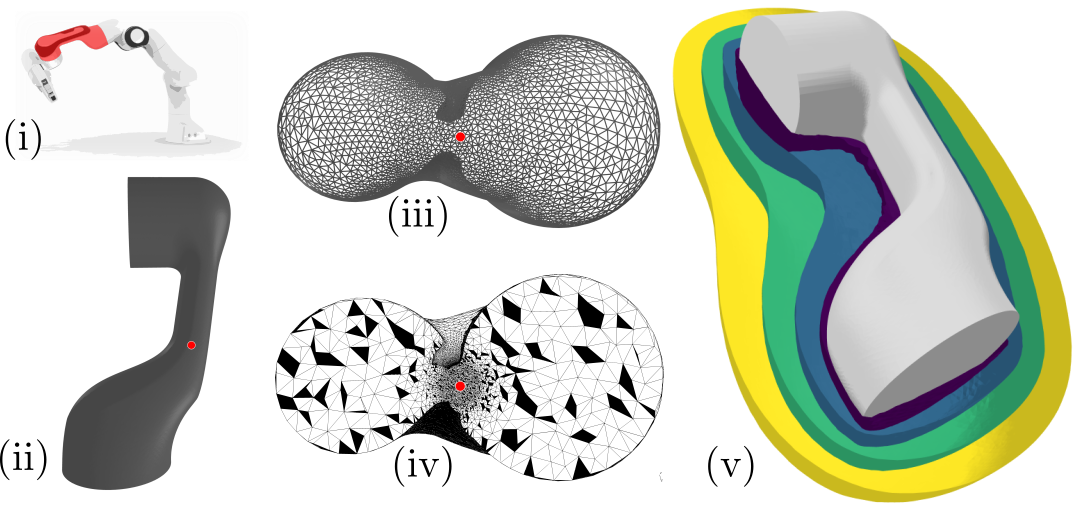}
    \caption{The computational method with the Kelvin Transformation approach is visualized in this figure, by using the 5th link of the Franka Robot (i) as an example object. The center of inversion is depicted as a red dot in insets (ii-iv). The inset (ii) shows the surface mesh of the object, (iii) shows the mesh transformed by the inversion map \eqref{eq:inversion_map}, (iv) shows the generated interior tetrahedral mesh, and (v) shows an iso-surface representation of the solution with constant boundary values. Corresponding numerical statistics are shown in Table \ref{tab:wrench_stats}.
    }
    \label{fig:method_link5}
\end{figure}

We introduce the notion of an object-wise \textit{semantics-aware distance function} that implicitly defines safe sets for contact-rich manipulation tasks under consideration of time-varying task semantics, such as object affordances, task goals or user intent.

In this problem statement, we define the desirable characteristics of the semantics-aware distance function and its corresponding computation method.
These characteristics should be met to ensure their usefulness in real-time control applications, such as ensuring the safety of actions commanded through teleoperation or learned policies.

We focus on the case of a rigid object with finite non-empty interior $\Theta \subset \mathbb{R}^3$ and surface $\partial\Theta \subset \mathbb{R}^2$. A possibly time-varying \textit{semantic state} $\sigma$ is defined on the object's surface $\sigma: \partial\Theta \to \mathbb{R}$, and represents whether contact at a point on the object's surface is allowed ($\sigma > 0$) or not ($\sigma \leq 0$).

Let a function $h:\mathbb{R}^3\setminus \Theta \to \mathbb{R}$ implicitly describe the semantic-aware safe set $\mathcal{C}$ as the superlevel set
\begin{align}
    \mathcal{C} &= \{\mathbf{x} \in \mathbb{R}^3 \setminus \Theta \ |\  h(\mathbf{x}) \geq 0\} \\
    \partial\mathcal{C} &= \{\mathbf{x} \in \mathbb{R}^3 \setminus \Theta \ |\  h(\mathbf{x}) = 0\},
\end{align}
borrowing notation from Ames et al. \cite{ames2019control}.
The function $h$ encodes the semantic state $\sigma$ as a scaled signed distance between the zero level set of $h$ and a point on the object's surface, i.e., 
\begin{align}
h(\mathbf{x}) = \gamma[d(\mathbf{x}, \partial\mathcal{C})],\quad \sigma(\mathbf{x}) - h(\mathbf{x}) = 0 \quad\mathbf{x} \in \partial\Theta,
\label{eq:h_cost}
\end{align}
where $\gamma$ is a class $\mathcal{K}$ function, hence, the absolute value of the semantic function $\sigma$ is directly related to the allowed contact velocity, or to the distance margin.
We call $h$ the \textit{semantics-aware distance function}, and aim to use it to constrain the motion of points in the object's exterior by the inequality constraint
\begin{align}
    \dot{h}(\mathbf{x}) \geq \alpha[ h(\mathbf{x})], \quad \mathbf{x} \in \mathbb{R}^3\setminus\Theta, \label{eq:h_cbf}
\end{align}
where $\alpha$ is a class $\mathcal{K}$ function. Assuming that $h$ is continuously differentiable in the domain $\mathbb{R}^3\setminus\Theta$, we use the function~$\alpha$ to ensure the forward-invariance of the safe set~$\mathcal{C}$~\cite{ames2019control}.

We require the following properties from the computation method: (i) the computed function should respect condition~\eqref{eq:h_cost}, (ii) the memory requirements should be on par to those of signed distance functions ($\sim 1\,\mathrm{MB}$) to adequately function in standard robotic systems \cite{li2023learning}, (iii) the query times $t_\mathrm{query}$ should be substantially lower than $\SI{1}{\milli\second}$ to achieve real-time control rates, and (iv) updates from a varying semantic state $\sigma$ should be considered online through update times \mbox{$t_\mathrm{solve} \leq \SI{30}{\milli\second}$}.

\section{Computation of Semantics-Aware Distances with the Kelvin Transformation} 
\label{sec:method}

 
As described in the preliminaries, an exact distance function can be obtained by applying the negative logarithm transformation \eqref{eq:transformation_varadhan} to the solution of a screened Poisson equation with Dirichlet boundary condition. We make use of this Partial differential equation (PDE) formulation to represent the semantics-aware distance in the condition \eqref{eq:h_cost}
\begin{align}
    \sigma - \lim_{c\to0} h_\partial = 0 \quad &\mathrm{on} \quad  \partial\Theta \\
    \sigma + \lim_{c\to0} \sqrt{c}\,\mathrm{ln}(u_\partial) = 0 \quad &\mathrm{on} \quad \partial\Theta \\
    u_\partial = \lim_{c\to0} \frac{1}{\sqrt{c}}\mathrm{exp}(-\sigma) \quad &\mathrm{on} \quad \partial\Theta.
\end{align}
Instead of solving the exact screened Poisson equation in the limit as $c\rightarrow0$, we propose to approximate the solution by the maximally regularized distance approximation, i.e., the solution of the Laplace equation with the vanishing condition at infinity
\begin{align}
    \Delta u = 0 \quad &\mathrm{in} \quad \Theta \notag \\
    u = \mathrm{exp}(-\sigma) \quad &\mathrm{on} \quad \partial\Theta \notag \\
    u \to 0 \quad &\mathrm{for} \quad |\mathbf{x}| \to \infty, \label{eq:ext_pde} 
\end{align}
benefitting from the fact that the regularized approximation is exact at the boundary, ensuring the satisfaction of condition~\eqref{eq:h_cost}. The solution is also ensured to be smooth and free of extrema in the domain interior~\cite{kim1992real}.

To solve the Laplace equation in unbounded domains, such as in the exterior of an object, we turn to the Kelvin Transformation approach and define the sets
\begin{align}
    \Omega := \mathbb{R}^3\setminus\Theta \cup \partial\Theta, \quad \partial\Omega := \partial\Theta.
\end{align}

\subsection{Solution with the Kelvin Transformation}
We follow the approach presented by Nabizadeh et al. to solve Laplace equations through the Kelvin Transformation approach \cite{nabizadeh2021kelvin}. The computational method is depicted in Fig. \ref{fig:method_link5}. The analytical behavior function $G(\mathbf{y})$ -- that captures the asymptotic behavior of the solution $u$ with the vanishing condition at infinity -- for Laplace equations is
\begin{align}
    G(\mathbf{y}) = |\mathbf{y}|, \quad \mathbf{y} \in \Omega_{\mathrm{inv}} \cup \{ \mathbf{0} \}.
\end{align}
Using the inversion map \eqref{eq:inversion_map} and the factorization \eqref{eq:kelvin_factorization}, the exterior boundary value problem \eqref{eq:ext_pde} can be exactly reformulated as
\begin{align}
    \Delta V(\mathbf{y}) = 0 \quad &\mathrm{in} \quad \Omega_\mathrm{inv} \cup \{ \mathbf{0} \} \notag \\
    V(\mathbf{y}) = V_\partial = \frac{U_\partial(\mathbf{y})}{G(\mathbf{y})} \quad &\mathrm{on} \quad \partial\Omega_\mathrm{inv} \label{eq:interior_pde}
\end{align}
in the now bounded and compactified domain $\Omega_\mathrm{inv} \cup \{ \mathbf{0} \}$, with $\Omega_\mathrm{inv} = \Phi^{-1}(\Omega)$ as presented in Section \ref{sec:prelim_kelvin}.

The safety inequality constraint \eqref{eq:h_cbf} requires the value and the time-derivative of the semantics-aware distance $h$, which is a function of the semantic state $\sigma$ and the evaluation point $\mathbf{x}$. Given that the evaluation is performed at orders of magnitude higher frequency than the semantic state adaptation, we approximate the time derivative
\begin{align}
    \dot{h} = \nabla_\mathbf{x}h \dot{\mathbf{x}} + \nabla_\sigma h \dot{\sigma} \approx \nabla_\mathbf{x}h \dot{\mathbf{x}} + \cancel{\nabla_\sigma h \dot{\sigma}}
\end{align}
by assuming a relative quasi-static behavior of the semantic state. The gradient $\nabla_\mathbf{x}h$ is continuously differentiable in the exterior domain $\Omega$ and can be computed by the chain rule.

\subsection{Linear Solver}

Based on a tetrahedral-mesh approximation (with $N_\mathrm{verts}$ vertices and $N_\mathrm{tets}$ tetrahedrons) of the bounded domain $\Omega_\mathrm{inv}\cup\{\mathbf{0}\}$, we represent the matrix form of the vertex-wise Laplace operator as $\mathbf{L} \in \mathbb{R}^{N_\mathrm{verts} \times N_\mathrm{verts}}$, and introduce the matrix $\mathbf{A} \in \mathbb{R}^{N_\mathrm{verts} \times N_\mathrm{verts}}$ as the coefficient matrix.

This results in the following linear PDE expressed in the bounded domain
\begin{align}
    \mathbf{A} V = \mathbf{0} \quad &\mathrm{in}  \quad \Omega_\mathrm{inv} \cup \{ \mathbf{0} \} \notag \\
    V = V_\partial \quad &\mathrm{on}  \quad \partial\Omega_\mathrm{inv}, \label{eq:interior_fem_pde}
\end{align}
with $\mathbf{A}=\mathbf{L}$ and $V$ as the vectorized vertex-wise solution.

The coefficient matrix $\mathbf{A} \in \mathbb{R}^{N_\mathrm{verts} \times N_\mathrm{verts}}$ is square and sparse. We reduce the linear system to solve only for the interior vertices, given that boundary values are known
\begin{align}
    \mathbf{b}_\Omega &= - \mathbf{A}_{\Omega, \partial}V_\partial \\
    \mathbf{A}_{\Omega, \Omega} V_\Omega &= \mathbf{b}_\Omega,
\end{align}
abusing notation with the $\Omega$ subscript representing the indices of vertices in the domain interior, and respectively $\partial$ analogue for vertices in the domain boundary.

The coefficient matrix can be factorized through an LU decomposition
\begin{align}
    \mathbf{L}_\mathbf{A}\mathbf{U}_\mathbf{A} = \mathbf{A}_\Omega,
\end{align}
to efficiently compute the solution with varying boundary values -- motivated by Crane et al. \cite{crane2013geodesics} -- enabling the desired adaptive behavior of the semantics-aware distance function
as encoded by the PDE solution.

\subsection{Spatial Discretization}
\label{sec:implementation}

We choose to discretize the bounded domain with a tetrahedral volumetric mesh, given that it leads to a piecewise smooth approximation of the object's surface and also allows for variable element sizing. We present the parametrization of the interior mesh generation, provide references to the definition of the differential operators employed, and describe the querying procedure.

\subsubsection{Parametric Tetrahedralization}
Ideally we want an interior mesh that is equivariant to the choice of inversion origin -- we propose and leverage the following metric based sizing field to generate the interior mesh. A uniform mesh in the exterior domain is obtained by meshing the interior with 
\begin{align}
    l_{\mathbf{y}, \mathrm{nom}}(\mathbf{y}) = |\mathbf{y}|^2 l_\mathbf{x}, \quad \mathbf{y} \in \Omega_\mathrm{inv}\cup\{ \mathbf{0} \}, \label{eq:ly_nom_sizing}
\end{align}
where $l_\mathbf{x}$ is the constant and nominal length-based element size in the exterior domain, and $l_\mathbf{y}(\mathbf{y})$ is the coordinate-dependent sizing field in the bounded domain.
These quantities are related by the Riemannian metric associated with the Kelvin Transformation \cite{nabizadeh2021kelvin}.

Note that the nominal interior sizing field \eqref{eq:ly_nom_sizing} leads to an infinite density at the origin in the bounded domain. 
We therefore set a lower bound for the sizing field, which depends on the original surface mesh. 

\subsubsection{Differential Operators}

To compute the vertex-wise Laplacian, we use the cotangent formula \cite{alexa2020properties, crane2019n}. 
To compute the vertex-wise gradient, we use the volume-weighted average gradient on the vertex's star. 
Due to the vertex-wise estimation, the resulting interpolated gradient within a tetrahedral mesh is continuously differentiable \cite{mancinelli2019comparison}.

\subsubsection{Value and Gradient Queries} \label{sec:query}

The procedure to query a value at a given point in the exterior domain is described in Algorithm \ref{alg:query}: ($1$) compute its inverse, ($2$) find the enclosing tetrahedron, ($3$) interpolate the vertex-wise interior solution $V$ from the enclosing tetrahedron, ($4$) multiply by the behavior function $G$, and  ($5$) transform through the negative logarithm map.

\begin{algorithm}[H]
    \caption{\small Query the Semantics-Aware Distance Function \eqref{eq:ext_pde}}
    \label{alg:query}
    \small
    \begin{algorithmic}[1]
        \Require Solution $V$ of interior problem \eqref{eq:interior_fem_pde}. Query point $\mathbf{x} \in \Omega$.
        \State $\mathbf{y} \gets \Phi^{-1}(\mathbf{x})$
        \State $\tau \gets \mathrm{locate}(\mathbf{y})$ \Comment{Find enclosing tetrahedron $\tau$}
        \State $V(\mathbf{y}) \gets \mathrm{barycentricInterpolation}(\mathbf{y}, \tau, V)$
        \State $U(\mathbf{y}) \gets G(\mathbf{y})V(\mathbf{y})$
        \State $h(\mathbf{x}) \gets -\mathrm{ln}[U(\mathbf{y})]$
        \State \Return $h(\mathbf{x})$
    \end{algorithmic}
\end{algorithm}

The gradients are obtained analogue to the value query algorithm, computing the gradient $\nabla_\mathbf{x}h(\mathbf{x})$ after the barycentric interpolation of the numerical gradients.

\section{Validation} 
\label{sec:validation}

In this section we present validation experiments on (i) the computational viability of the method for real applications, and (ii) a demonstration on a complex object with various semantic regions.

\subsection{Numerical Evaluation} \label{sec:numerical_eval}

In this experiment, we aim to quantitatively assess whether the proposed method fulfills the requirements and desired characteristics as described in Sec. \ref{sec:problem_statement}. 

\begin{figure}
    \centering
    \includegraphics[width=\linewidth]{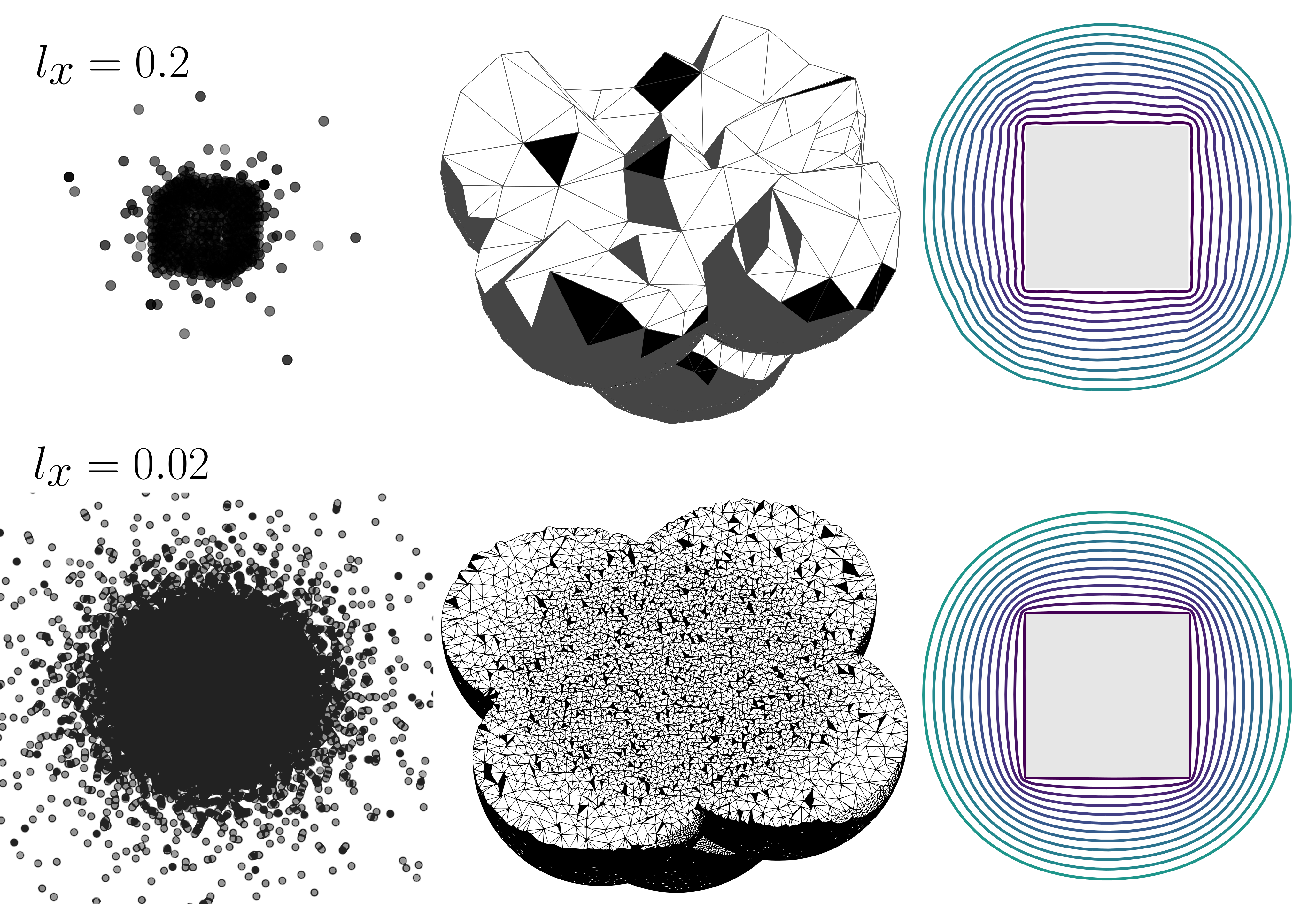}
    \caption{Visualization of the tested mesh range, as parametrized by the exterior domain length $l_x$. Top row: $l_x=0.2$, bottom row: $l_x=0.02$. The vertices are projected to the exterior domain through the inversion map~\eqref{eq:inversion_map}, and shown in the left column. The mesh of the inverse domain is shown in the center. An iso-lines representation of the function evaluated in a grid of size $(1\,000 \times 1\,000)$ within the cube $[-3, 3]$ is shown in the right column.}
    \label{fig:mesh_range}
\end{figure}

\begin{table*}[t]
    \centering
    \caption{Comparison of Size, Quality, and Computation time, evaluated as a function of varying mesh resolution. The mesh resolution is parametrized by the exterior length $l_x$, described in Section \ref{sec:implementation}, and depicted in Fig. \ref{fig:mesh_range}. The reported times for mesh,  build, and decompose, are one-point estimates. The reported solve time is the mean of 100 evaluations. The reported query time is the median of a thousand points and using a cell hint. The RMSE is computed using the solution at 1M points in a planar grid and wrt. the solution obtained at $l_x=0.02$.}
    \begin{tabular}{|c|c|c|c|c|c|c|c|c|c|}
    \hline
    \rule{0pt}{10pt}
    $l_x$ & $N_\mathrm{verts}$ & $N_\mathrm{tets}$ & Size$/ \mathrm{MB}$ & $\mathrm{RMSE}_{l_x = 0.02}$ & $t_\mathrm{mesh} / \SI{}{\second}$ & $t_\mathrm{build} / \SI{}{\second}$  & $t_\mathrm{decomp} / \SI{}{\second} $ & $t_\mathrm{solve} / \SI{}{\milli\second}$ & $t_\mathrm{query} / \SI{}{\micro\second}$\\[3pt]
    \hline
    &&&&&&&&&\\[-0.8em] 
    $ 0.02 $ & $ 157\,448 $ & $ 864\,083 $ & $ 38.0 $ & $ -     $ & $ 12.4   $ & $ 1545  $ & $ 685.7 $ & $ 559. \pm 4.54  $  & $ 0.570 $ \\
    $ 0.03 $ & $ 52\,865  $ & $ 273\,407 $ & $ 12.1 $ & $ 0.002 $ & $ 4.00   $ & $ 209.6 $ & $ 55.12 $ & $ 97.5 \pm 1.18 $  & $ 0.414 $ \\
    $ 0.04 $ & $ 24\,786  $ & $ 121\,965 $ & $ 5.6  $ & $ 0.004 $ & $ 1.82   $ & $ 43.32 $ & $ 9.140 $ & $ 28.6 \pm 0.61 $  & $ 0.283 $ \\
    $ 0.05 $ & $ 14\,124  $ & $ 66\,402  $ & $ 3.1  $ & $ 0.006 $ & $ 1.05   $ & $ 12.83 $ & $ 2.112 $ & $ 10.6 \pm 0.37 $  & $ 0.279 $ \\
    $ 0.06 $ & $ 8\,910   $ & $ 40\,610  $ & $ 1.8  $ & $ 0.009 $ & $ 0.68   $ & $ 4.256 $ & $ 0.801 $ & $ 5.31 \pm 0.36 $  & $ 0.251 $ \\
    $ 0.08 $ & $ 4\,415   $ & $ 18\,723  $ & $ 0.9  $ & $ 0.013 $ & $ 0.34   $ & $ 0.854 $ & $ 0.122 $ & $ 1.50 \pm 0.21 $  & $ 0.217 $ \\
    $ 0.10  $ & $ 2\,751   $ & $ 11\,109  $ & $ 0.5  $ & $ 0.018 $ & $  0.22  $ & $ 0.333 $ & $ 0.048 $ & $ 0.63 \pm 0.15 $  & $ 0.175 $ \\
    $ 0.15 $ & $ 1\,132   $ & $ 4\,125   $ & $ 0.2  $ & $ 0.032 $ & $  0.84  $ & $ 0.075 $ & $ 0.011 $ & $ 0.13 \pm 0.04 $  & $ 0.211 $ \\
    $ 0.20  $ & $ 615    $ & $ 2\,030   $ & $ 0.1  $ & $ 0.046 $ & $  0.48  $ & $ 0.015 $ & $ 0.004 $ & $ 0.05 \pm 0.01 $  & $ 0.116 $ \\
    \hline
    \end{tabular}%
    \label{tab:stats}
\end{table*}

\begin{figure}
    \centering
    \includegraphics[width=\linewidth]{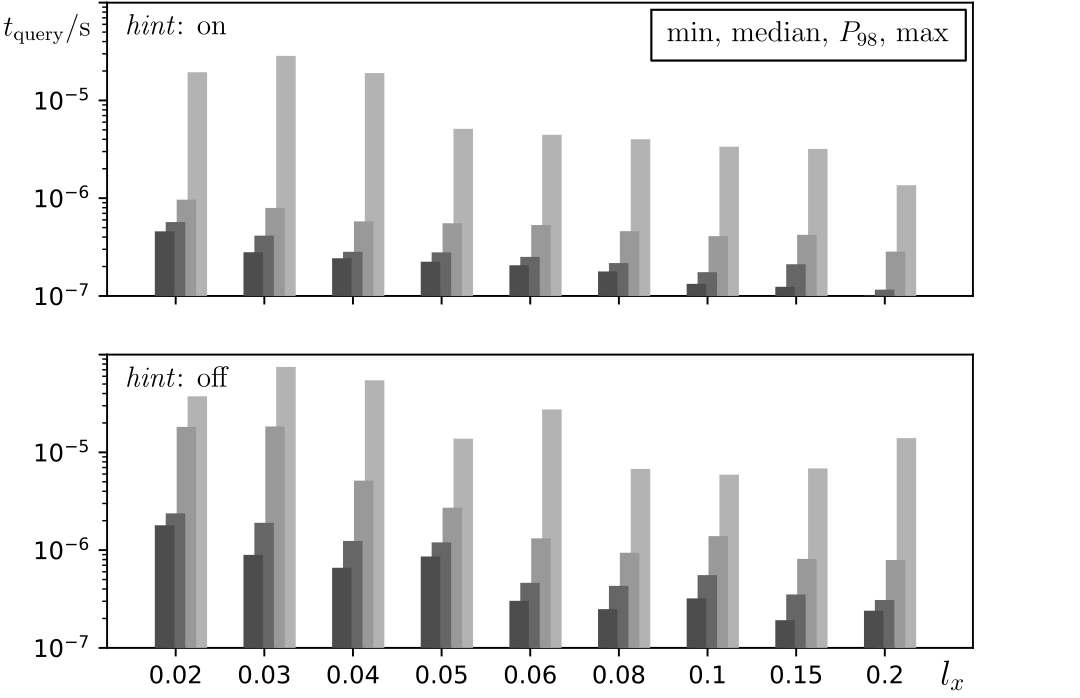}
    \caption{Comparison of query time statistics as a function of the mesh resolution parameter $l_x$, and the usage of a hint (warm start). In continuous motion, the hint can be set as the previous result (enclosing cell or tetra), and leads to query times considerably under a microsecond.}
    \label{fig:query_times}
\end{figure}

Given that our method is based on a numerical solution of the PDE \eqref{eq:ext_pde}, we measure its computational properties as a function of mesh resolution. 
We vary the parameter $l_x$ that determines the mesh generation process in the inverse or bounded domain, and measure method-characterizing size, quality, and performance related variables.

In this evaluation we use the cube as an example shape, leading to intuitive parametrizations, simple visualizations, and to study the behavior of the method in the presence of sharp edges. The inversion origin is placed at the center of the cube. All computations were performed on a single core (i7-8750H CPU @ $\SI{2.20}{\GHz}$), implemented in \textit{C++}, using the (standard) supernodal sparse LU factorization in \textit{Eigen} \cite{eigenweb}. The mesh generation and \textit{locate} queries make use of the open source \textit{CGAL} \cite{cgal:eb-24a} implementations.

Fig.~\ref{fig:mesh_range} shows the range of mesh resolutions through a subset of the vertices projected to the exterior domain, the generated interior mesh, and the iso-lines of the resulting functions $h(u_{l_x})$, evaluated with boundary values $u_{\partial} = 1$. 

\vspace{10pt}

Two expected behaviors become clear from the iso-line visualizations: 
\begin{enumerate}
    \item the numerical solution becomes smoother as the mesh becomes finer, and
    \item the iso-lines approach the geometry of the boundary exactly as $h(u) \rightarrow 0$ for all evaluated mesh resolutions.
\end{enumerate}
In addition to these qualitative observations, the numerical results of this evaluation are presented in Table \ref{tab:stats} and Fig. \ref{fig:query_times}.
The sizes of the evaluated models lie in the range ($0.1 -  38\,\mathrm{MB}$), which can be effectively managed by modern computing platforms in robotic systems, thanks to the recent increases in memory availability and efficiency.
The quality of the solution should increase, as $l_x$ decreases. We evaluate and corroborate this expectation through a proxy measurement using the RMSE with respect to the solution computed at $l_x=0.02$.

The times for (i) 3D mesh generation $t_\mathrm{mesh}$, (ii) constructing the sparse Laplace matrix operator $t_\mathrm{build}$, and (iii) computing the LU decomposition $t_\mathrm{decomp}$ are also shown in Table \ref{tab:stats}. As stated above, all computations were performed using a single CPU core, meaning that the time dedicated to preprocessing steps could be considerably reduced, e.g., by leveraging parallel algorithms to construct the Laplace operator.

The solving time is the key factor in determining whether the method supports online-adaptive behavior.
The solving times in the evaluated meshes lie in the range ($0.05 - \SI{559}{\milli\second}$), and are sufficiently low to enable online-recomputations of the solution. We assume that camera-based updates related to task level changes are streamed at a $\SI{30}{\Hz}$ frequency, and could therefore be processed by the tested models up to (incl.) $l_x=0.04$.

With CGAL tetrahedralizations, we can use a hint or warm-start the \textit{locate} query to find the enclosing tetrahedron given a point in the discretized domain. The effect of this hint on the query times is shown in Fig. \ref{fig:query_times}. With and without the hint, the query times are more-than-sufficiently low ($\SI{0.5}{\micro\second}$) for real-time applications. The usage of a hint leads to considerably faster queries, and is especially simple to leverage in continuous motion or point tracking problems. Because of the triangulation-walking nature of the query algorithm, relatively large query times can arise ($\SI{10}{\micro\second}$), yet seldom, as shown in Fig. \ref{fig:query_times}.

\subsection{Evaluation with varying Semantic-State}

In this validation experiment we demonstrate the semantic-dependency of our method on a combination wrench as a 3-dimensional example.
To focus on the analysis of the method, we manually design three different semantic scenarios for the tool. However, note that pretrained models for affordance-detection such as \cite{geng2023rlafford, tabib2024lgafford} can replace or enhance the manual design and lead to more practical implementations.

We evaluate our method on a combination wrench, to (i) assess the performance of the method on a thin object that is also topologically distinct from a sphere, and (ii) visualize the encoding of semantics that are relevant to tool-based manipulation tasks. Given that the method is agnostic to the object's geometry, we expect the results to directly generalize without any further changes in the parameters, apart from the object-specific mesh generation.

The wrench has a length of $\SI{96.47}{\milli\meter}$. The length $l_{x,\mathrm{wrench}}$ parametrizing the mesh generation is set to $\SI{0.37}{\milli\meter}$ to accurately represent the surface geometry. Further, we set the inversion origin at the center of the wrench.

The numerical evaluation is shown in Table \ref{tab:wrench_stats}. 
The performance lies within the expected ranges observed in the cube-based numerical evaluation in Section \ref{sec:numerical_eval}.
We interpret this observation as a strong empirical support for the geometry-independent behavior of the method.

\begin{figure}
    \centering
    \includegraphics[width=\linewidth]{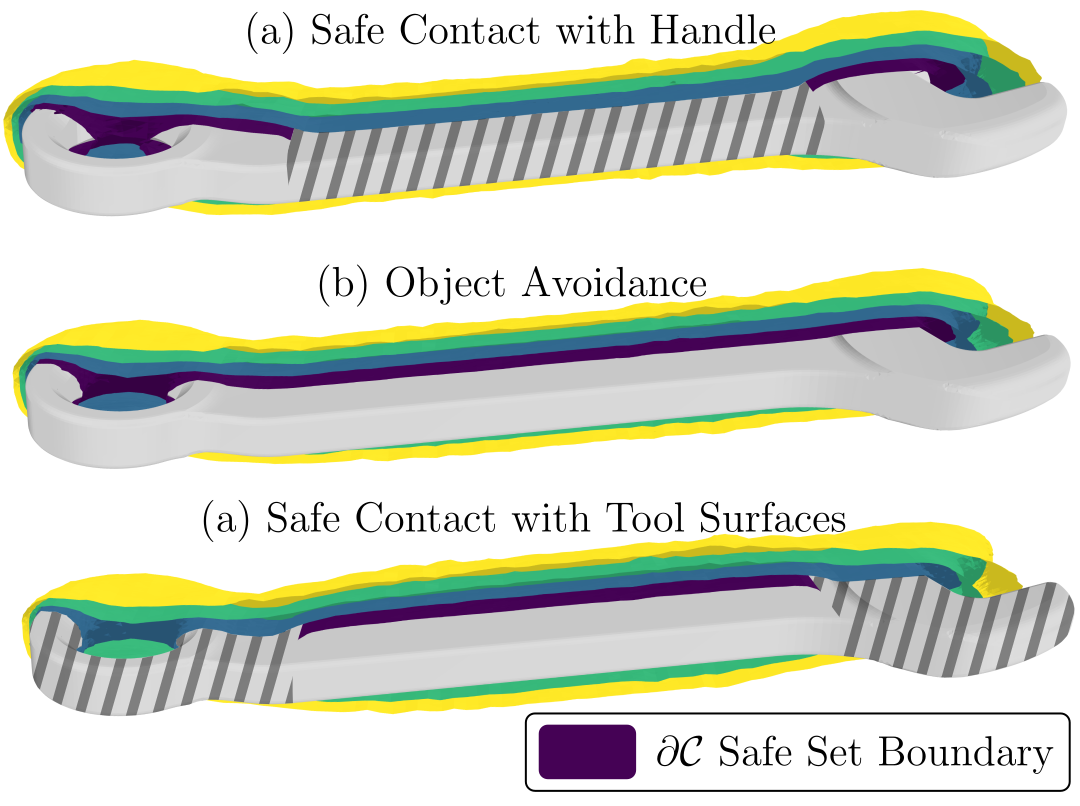}
    \vspace{-15pt}
    \caption{We visualize the semantics-aware distance of a combination wrench through its iso-surfaces, at three different semantic configurations. The striped surface regions represent the surface regions where contact is allowed, i.e., where the zero level-set of the computed function lies within the object.}
    \label{fig:main_wrenches}
\end{figure}

\begin{table}
\centering
\caption{Numerical Results of the Combination Wrench Evaluation. The columns are a subset of the ones in Table \ref{tab:stats} and share their description.}
\begin{tabular}{|c|c|c|}
\hline
\rule{0pt}{10pt}
Parameter & Comb. Wrench (Fig.\ref{fig:main_wrenches}) & Franka Link 5 (Fig. \ref{fig:method_link5}) \\
\hline
\rule{0pt}{10pt}
$N_\mathrm{verts}$ & $30\,749$ & $37\,294$ \\
$N_\mathrm{tets}$ & $141\,546$ & $215\,736$ \\
Size$/ \mathrm{MB}$ & $7.2$ & $8.4$ \\
$t_\mathrm{mesh} / \SI{}{\second}$ & $3.68$ & $2.201$ \\
$t_\mathrm{build} / \SI{}{\second}$ & $69.03$ & $100.2$ \\
$t_\mathrm{decomp} / \SI{}{\second}$ & $2.16$ & $33.46$\\
$t_\mathrm{solve} / \SI{}{\milli\second}$ & $14.6 \pm 0.56$ & $67.7 \pm 1.32$ \\
$t_\mathrm{query} / \SI{}{\micro\second}$ & $0.324$ & $0.301$ \\[0.3em]
\hline
\end{tabular}
\label{tab:wrench_stats}
\end{table}

Regarding the applications and varying semantics, we consider multiple usages of the tool depending on the allowed contact surfaces: (a) only contact with the handle is allowed, (b) any contact with the object should be avoided, and (c) only contact with the tool ends or functional surfaces is allowed. 
To this end, we parametrize the semantic state~$\sigma(\mathbf{x})$ to be dependent on the point's coordinate $z$ along the longitudinal axis $\mathbf{e}_l$ of the combination wrench. 

Concretely,
\begin{align}
    \sigma(\mathbf{x}) = \mathrm{tanh}(\mathrm{abs}(\mathbf{e}_l^T\mathbf{x})-d_0))\sigma_\mathrm{nom},
\end{align}
where $\sigma_\mathrm{nom}$ is a nominal constant semantic value, and $d_0 \in \mathbb{R}_{\geq0}$ represents the symmetric location of the transition between semantic regions. We implement the cases (a)-(c) listed above
The resulting functions are visualized through iso-surfaces in Fig. \ref{fig:main_wrenches}, and match the designed safe sets.

\section{Discussion and Conclusion} 
\label{sec:conclusion}

We introduce the \textit{semantics-aware distance} and propose a numerical method to compute it, enabling safe reactive behavior while adaptively considering regional contact constraints in contact-rich manipulation tasks. 

Our method generalizes the applicability of safety-related methods in robotic manipulation from collision avoidance towards the consideration of object affordances and semantics while performing and learning contact-rich manipulation.

While the semantics-aware distance is independent of the computation method, our approach assumes that objects have a known geometry and are provided as mesh-based representations.
Our method applies therefore to interactions with known objects, which are ubiquitous in robotic manipulation tasks. 
Further, although the solutions of the Laplace equation are smooth in the domain interior, our finite-element discretization results instead in piecewise-smooth solutions. 

These two limitations lead to interesting and relevant extensions of the method: developing the method to consider unknown objects along the lines of \cite{maric2024online}, and approximating solutions through analytical basis functions without sacrificing accuracy at the boundary.

The sub-microsecond query times and the continuously differentiable gradients of the function motivate applications within planning and trajectory optimization for whole-body manipulation tasks, not only from a safety perspective but also by leveraging harmonic potential fields for globally-informed guidance. A further related application is semantics-aware safety critical control during teleoperation or kinesthetic teaching of robots in open environments, especially in combination with available pretrained affordance-detection models. 

\vspace{3em}


\balance
\bibliographystyle{unsrt}
\bibliography{references}
\end{document}